\documentclass[letterpaper, 10 pt, conference]{ieeeconf}  

\IEEEoverridecommandlockouts                              

\usepackage{graphics} 
\usepackage{graphicx}

\usepackage{epsfig} 
\usepackage{amsmath} 
\usepackage{amssymb}  

\usepackage{bm}
\usepackage{mathtools}

\usepackage{algorithm}
\usepackage[noend]{algpseudocode} 
\algnewcommand\AAND{\textbf{ and }}
\algnewcommand\Or{\textbf{ or }}

\usepackage{color}
\usepackage{citesort}
\usepackage{flushend}
\usepackage{url}
\usepackage[breaklinks]{hyperref}
\usepackage[capitalise]{cleveref}
\usepackage{booktabs}
\usepackage{makecell}
\usepackage{multirow}
\usepackage{siunitx}
\usepackage[nolist,nohyperlinks]{acronym}
\acrodef{method}[AOM]{ACRONYM OF METHOD}
\acrodef{gnss}[GNSS]{Global Navigation Satellite System}
\acrodef{ransac}[RANSAC]{Random Sample Consensus}
\acrodef{slam}[SLAM]{Simultaneous Localization And Mapping}
\acrodef{pca}[PCA]{Principal Component Analysis}
\acrodef{ekf}[EKF]{Extended Kalman Filter}
\acrodef{rmse}[RMSE]{Root Mean Square Error} 
\acrodef{ape}[APE]{Absolute Pose Error}
\acrodef{cfar}[CFAR]{Constant False Alarm Rate}
\acrodef{snr}[SNR]{Signal to Noise Ratio}
\acrodef{rcs}[RCS]{Radar Cross Section}
\acrodef{imu}[IMU]{Inertial Measurement Unit}
\acrodef{sgm}[SGM]{Segmi-Global Matching}
\acrodef{dnn}[DNN]{Deep Neural Network}
\acrodef{gru}[GRU]{Gated Recurrent Unit}
\acrodef{hpr}[HPR]{Hidden Point Removal}
\acrodef{raft}[RAFT]{Recurrent All-Pairs Field Transforms}
\acrodef{fov}[FOV]{Field of View}
\acrodef{mclab}[MC-lab]{Marine Cybernetics laboratory}
\acrodef{vio}[VIO]{Visual-Inertial Odometry}
\acrodef{rcm}[RCM]{Refractive Camera Model}
\acrodef{sfm}[SFM]{Structure from Motion}
\acrodef{cnn}[CNN]{Convolutional Neural Network}
\acrodef{mse}[MSE]{Mean Squared Error}
\acrodef{gnll}[NLL]{Negative Log Likelihood}
\acrodef{rov}[ROV]{Remotely Operated Vehicle}
\acrodef{deepvl}[DeepVL]{Deep Velocity Learning}

\acrodef{rpe}[RPE]{Relative Position Error}
\acrodef{dvl}[DVL]{Doppler Velocity Log}
\acrodef{reaqrovio}[ReAqROVIO]{Refractive Aquatic ROVIO}

\newcommand{\vel}{\mathbf{v}}
\newcommand{\velpred}{\mathbf{\hat{v}}}
\newcommand{\velunc}{\mathbf{\hat{u}}}
\newcommand{\sigmapred}{\mathbf{\hat{\Sigma}}}
\newcommand{\accmeas}{\tilde{\mathbf{a}}_{\bodyframe}}
\newcommand{\gyromeas}{\tilde{\boldsymbol{\omega}}_{\bodyframe}}

\newcommand{\accbody}{\mathbf{a}_{\bodyframe}}
\newcommand{\gyrobody}{\boldsymbol{\omega}_{\bodyframe}}
\newcommand{\bodyframe}{\mathcal{B}}   
\usepackage{amsmath}
\usepackage{tabularray}

\usepackage{tikz}
\usetikzlibrary{positioning, shapes.geometric, arrows.meta, decorations.pathreplacing, calligraphy}

\DeclareMathAlphabet{\pazocal}{OMS}{zplm}{m}{n}

\DeclareMathAlphabet{\mathpzc}{OT1}{pzc}{m}{it}

\usepackage{array}
\newcolumntype{C}[1]{>{\centering\arraybackslash}p{#1}}
\newcolumntype{M}[1]{>{\raggedright\arraybackslash}p{#1}}

\usepackage{array} 
\newcolumntype{L}[1]{>{\raggedright\let\newline\\\arraybackslash\hspace{0pt}}m{#1}}	
\newcolumntype{S}[1]{>{\centering\let\newline\\\arraybackslash\hspace{0pt}}m{#1}}
\newcolumntype{R}[1]{>{\raggedleft\let\newline\\\arraybackslash\hspace{0pt}}m{#1}}

\makeatletter
\renewcommand*{\@opargbegintheorem}[3]{\trivlist
  \item[\hskip \labelsep{\itshape #1\ #2}] \textit{(#3)}\ }
\makeatother

\title{\LARGE \bf
DeepVL: Dynamics and Inertial Measurements-based Deep Velocity Learning for Underwater Odometry 
}

\author{Mohit Singh, and Kostas Alexis 
\thanks{This material was supported by the Research Council of Norway Award NO-327292.}
\thanks{The authors are with the Norwegian University of Science and Technology (NTNU), O. S. Bragstads Plass 2D, 7034, Trondheim, Norway {\tt\small mohit.singh@ntnu.no}}
}

\begin{document}

\maketitle
\thispagestyle{empty}
\pagestyle{empty}

\begin{abstract}
This paper presents a learned model to predict the robot-centric velocity of an underwater robot through dynamics-aware proprioception. The method exploits a recurrent neural network using as inputs inertial cues, motor commands, and battery voltage readings alongside the hidden state of the previous time-step to output robust velocity estimates and their associated uncertainty. An ensemble of networks is utilized to enhance the velocity and uncertainty predictions. Fusing the network's outputs into an Extended Kalman Filter, alongside inertial predictions and barometer updates, the method enables long-term underwater odometry without further exteroception. Furthermore, when integrated into visual-inertial odometry, the method assists in enhanced estimation resilience when dealing with an order of magnitude fewer total features tracked (as few as $1$) as compared to conventional visual-inertial systems. Tested onboard an underwater robot deployed both in a laboratory pool and the Trondheim Fjord, the method takes less than $5\textrm{ms}$ for inference either on the CPU or the GPU of an NVIDIA Orin AGX and demonstrates less than \SI{4}{\percent} relative position error in novel trajectories during complete visual blackout, and approximately \SI{2}{\percent} relative error when a maximum of $2$ visual features from a monocular camera are available. 

\end{abstract}

\section{INTRODUCTION}
Resilience is a crucial aspect of expanding the envelope of robotics. In the underwater domain, a core challenge is achieving long-term reliable state estimation. Traditionally, underwater systems depend exceedingly on high-end acoustic sensing alongside specialized \acp{imu}~\cite{wu2019survey} to ensure the availability of high-quality data supporting state estimation. On the other hand, a recent body of works exploring camera-based underwater navigation is challenged when it comes to dealing with the many intricacies of underwater vision, including low-light, poor visibility, refractive effects, and more~\cite{singh2024online,miao2021univio,ferrera2019aqualoc,gu2019environment}. Strategies for enhancing the resilience of odometry estimation are therefore essential.

\begin{figure}
    \centering
    \includegraphics[width=0.99\linewidth]{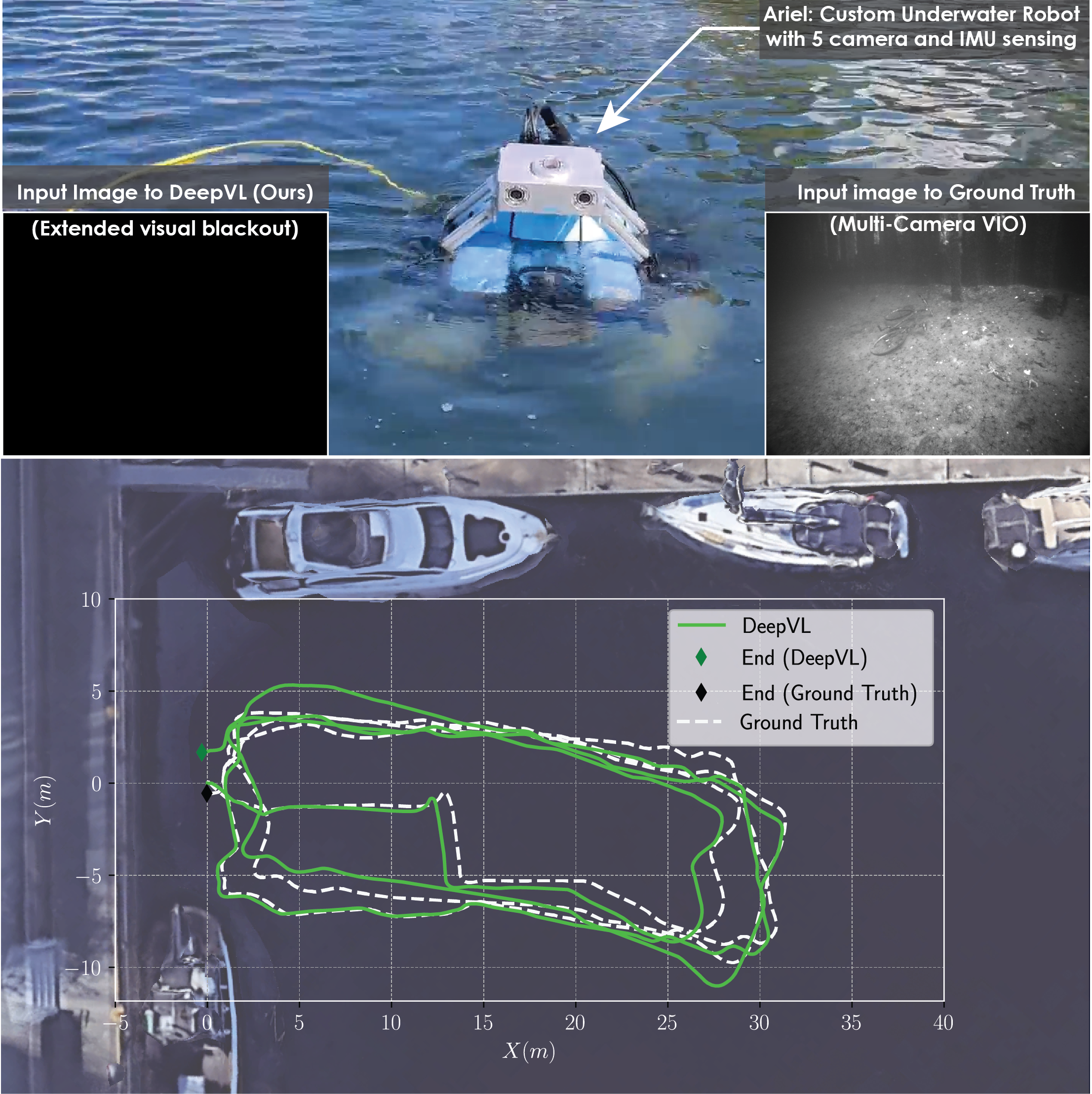}
    \vspace{-4ex}
    \caption{Utilized custom underwater robot with 5 camera visual-inertial sensing alongside overlay of estimated trajectory from the proposed method and ground-truth \ac{vio} on a top-down view of a pier in the Trondheim Fjord.}
    \vspace{-4ex}
    \label{fig:introfigure}
\end{figure}\par
In response, we present \ac{deepvl} a novel method for long-term reliable state estimation in challenging real-world scenarios based on a dynamics-aware neural odometry solution to enable localization in visual blackout, alongside assisting vision-based odometry subject to extreme lack of visual features (i.e., as few as $1$ feature). We propose a deep recurrent neural network-based model trained on sequences of proprioceptive robot data to predict the instantaneous robot-centric velocity and corresponding uncertainty. Particularly, we use both \ac{imu} measurements and the thruster command data driven by the complementary information the two modalities offer about the robot dynamics, alongside battery level readings. When fused as a velocity update to an \ac{ekf} with \ac{imu}-based prediction and barometer updates, the method offers persistent odometry without other exteroception, while when integrated into \ac{vio} it allows similar odometry results with an order of magnitude fewer features tracked as compared to conventional \ac{vio}. The \ac{deepvl} network is small and can seamlessly run on CPU, thus allowing it to be deployed on the computing solutions of existing underwater robots without modifications. Likewise, the method does not assume high-quality \ac{imu} data.

The proposed neural model is trained predominantly on data from an indoor laboratory alongside missions in the Trondheim Fjord (Figure \ref{fig:introfigure}), while a set of distinct trajectories are used for its evaluation. Specifically, the method was evaluated on $6$ trajectories from the Trondheim Fjord and $2$ in an indoor pool with varying motions and trajectory lengths (\SI{100}{\meter} to \SI{300}{\meter}). First, we evaluate the fusion of \ac{deepvl} in an \ac{ekf} with \ac{imu} propagation and barometer readings, and a detailed analysis is presented. Secondly, the effect of fusing visual features (ranging from $1$ to $8$), to the estimator is studied in an extensive ablative study. Third, we evaluate the consistency of the method's uncertainty estimation. Lastly, we present a validation of way-point tracking with state estimation feedback provided by the proposed method. As demonstrated, the presented work allows odometry with \SI{3.9}{\percent} relative error for extended continuous visual blackout and \SI{2.2}{\percent} for a maximum of $2$ visual features from a monocular camera. 

In the remainder of this paper, Section~\ref{sec:relatedwork} presents related work, while the deep velocity learning model is detailed in Section~\ref{sec:deepvl} and its integration in estimation is presented in Section~\ref{sec:vio}. Experimental studies are shown in Section~\ref{sec:evaluation}, while conclusions are drawn in Section~\ref{sec:concl}.

\section{RELATED WORK}\label{sec:relatedwork}
Prediction of odometry on the basis of \acp{imu} has a long history in the military and civilian domains alike~\cite{titterton2004strapdown}. Considering the explicit goal of inertial-only estimation of odometry for pedestrians with \acp{imu} integrated on the shoes, \cite{angermann2012footslam} utilized a Bayesian approach to achieve long-term error stability. Exploiting the power of \acp{dnn}, the work in~\cite{ionet} considers segments of inertial data and achieves to estimate non-periodic motion trajectories. Focusing on pedestrians carrying a smartphone and performing natural motions, the work in~\cite{ronin} explored different \ac{dnn} architectures and demonstrated robust performance. TLIO~\cite{liuTLIOTightLearned2020a} built on top of such ideas and proposed a tightly-coupled \ac{ekf}-based method for odometry estimation relying exclusively on inertial sensor data. 

Currently, the domain of exclusively \ac{imu}-based odometry estimation has developed into a diverse field including a multitude of methods deployed across diverse robot configurations~\cite{cohen2023inertial}. The work in~\cite{cioffiLearnedInertialOdometry2023a} successfully deployed learned inertial odometry in the framework of autonomous drone racing. Considering neural prediction as part of an on-manifold \ac{ekf} estimator, the method in~\cite{bajwa2024dive} presented low prediction errors for quadrotor trajectories. For ground vehicles, the authors in~\cite{karlsson2021speed} presented speed estimation based on a \ac{cnn} with accelerometer and gyroscope measurements as inputs. This line of works has extended to legged robots~\cite{buchananLearningInertialOdometry2022}, as well as underwater systems~\cite{liDIEMEDeepInertial2024}. Methodologically, the literature presents diversity both regarding the neural architectures employed~\cite{chen2024deep} and in terms of using different estimation techniques, e.g., factor graphs~\cite{buchananDeepIMUBias2023a}. 

However, reliance exclusively on inertial sensing presents significant limitations especially when it comes to long-term consistency. At the same time, complementary proprioceptive signals may be seamlessly available onboard a robotic system raising the potential for multi-modal approaches to improve resilience. To that end, the work in~\cite{zhangDIDODeepInertial2022} proposed Deep Inertial quadrotor Dynamical Odometry (DIDO) which employs \acp{dnn} to learn both the IMU and dynamics properties and thus better support state estimation. Key to its approach is the use of tachometer sensing onboard the quadrotor's motors. From a different perspective, the authors in~\cite{joshiSMVIORobust2023a} consider underwater \ac{vio} combined with a model-based motion estimator such that when \ac{vio}-alone fails the collective system presents robustness. 

Compared to current literature, our contributions include: \textit{i)} a dynamics-aware proprioceptive method that accurately predicts an underwater robot's velocities and associated covariances through the fusion of \ac{imu} and motor information, \textit{ii)} an implementation based on single input-single output with the temporal context being propagated over the hidden states of the \ac{gru}, significantly reducing the computational overhead compared to a buffer of inputs, \textit{iii)} use of readily available motor commands instead of specialised RPM sensing, \textit{iv)} a network learning the velocity prediction with about $28$k parameters which allows for fast inference on both CPU and GPU, \textit{v)} the use of an ensemble of networks to enhance velocity and uncertainty prediction, and \textit{vi)} flexible fusion of the method into odometry estimation either without or with visual updates. 

\section{Deep Velocity Learning}\label{sec:deepvl}

\subsection{Proprioceptive Inputs}
\subsubsection{Inertial Measurement Unit} The \ac{imu} is assumed to be rigidly attached to the robot, the robot body frame $\bodyframe$ is defined as the \ac{imu} coordinate frame, and the world frame is defined as $\mathcal{W}$. The $\ac{imu}$ provides the 3D accelerometer measurements $\accmeas$ and 3D gyroscope measurements $\gyromeas$ which are modelled as:
\begin{subequations}
\begin{align}
    \accmeas&=\accbody + \mathbf{b}_{\mathbf{a}} + \mathbf{q}^{-1}\mathbf{g}_{\mathcal{W}} + \mathbf{n}_{\mathbf{a}}\\
    \gyromeas&=\gyrobody + \mathbf{b}_{\boldsymbol{\omega}} + \mathbf{n}_{\boldsymbol{\omega}}
\end{align}
\end{subequations}
where $\accbody$ is the linear acceleration of the robot in the frame $\bodyframe$, $\mathbf{b}_\mathbf{a}$ is the acceleration bias, modelled as a random walk $\dot{\mathbf{b}_\mathbf{a}}\sim\mathcal{N}(\mathbf{0}, \Sigma_{\mathbf{b}_{\mathbf{a}}})$ and $\mathbf{n}_{\mathbf{a}}$ is the noise $\mathbf{n}_{\mathbf{a}}\sim\mathcal{N}\left(\boldsymbol{0},\Sigma_{\mathbf{a}}\right)$. The orientation of the robot is defined by $\mathbf{q}$ as a map from $\mathcal{B}$ to $\mathcal{W}$ and $\mathbf{g}_{\mathcal{W}}$ is the gravity vector aligned with $Z$ axis of the frame $\mathcal{W}$. Similarly, $\gyrobody$ denotes the angular velocity of the robot in $\bodyframe$, $\mathbf{b}_{\boldsymbol{\omega}}$ denotes the gyroscope bias modelled as a random walk $\dot{\mathbf{b}_{\boldsymbol{\omega}}}\sim\mathcal{N}\left(\mathbf{0}, \Sigma_{\mathbf{b}_{\boldsymbol{\omega}}}\right)$ and $\mathbf{n}_{\boldsymbol{\omega}}$ denotes the noise $\mathbf{n}_{\boldsymbol{\omega}}\sim\mathcal{N}\left(\mathbf{0}, \Sigma_{\boldsymbol{\omega}}\right)$.

\subsubsection{Motor Commands} Given a robot consisting of a total of $J$ rigidly attached bidirectional thrusters, whose input motor commands are $\left\{u_{j}: j\in\left[1, J\right]\right\}$, the combined thruster commands are expressed as a vector $\mathbf{u}_{J}=\left[u_{1}, u_{2},..., u_{J}\right]$. In the proposed method we use the actuator commands from the onboard autopilot.
\subsubsection{Battery Voltage}
The battery voltage is used to mitigate the dependence of thrust generated by the robot and the input motor command. The battery input is denoted as $k_{v}$ which is the instantaneous voltage of the battery.
\subsubsection{Network Input}
The input modalities described above are stacked into a $7+J$ channel vector at every time step $t$.
\begin{equation}
\mathbf{p}_{t}=\left[\accmeas,\gyromeas, \mathbf{u}_{J}, k_{v}\right]_{t}
\end{equation}

\begin{figure}
    \centering
    \includegraphics[width=1.0\linewidth]{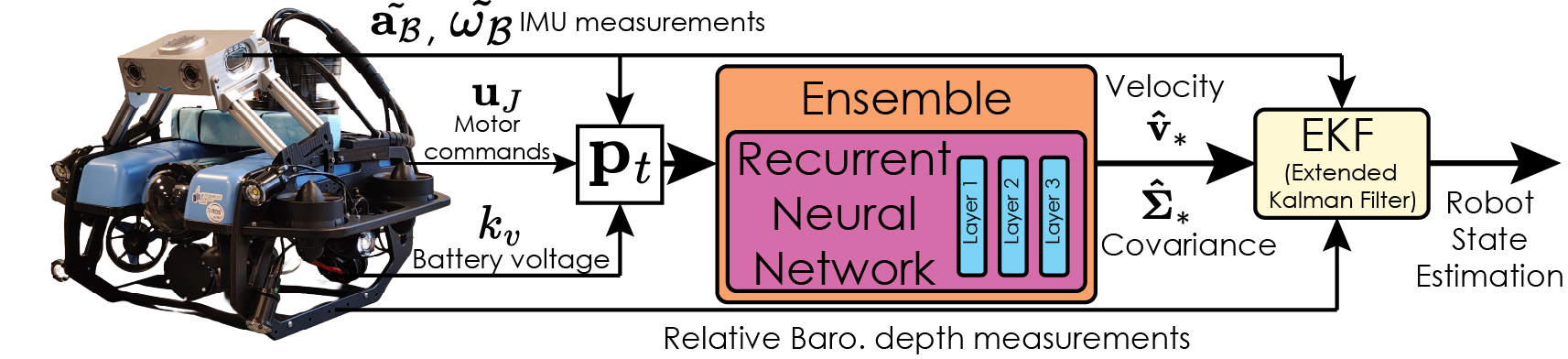}
    \caption{DeepVL method overview with \ac{imu}, motor commands and battery voltage as proprioceptive inputs to the ensemble of recurrent neural networks. The output velocity and covariance alongside the relative barometric depth measurement are then used in an \ac{ekf} for robot state estimation.}
    \label{fig:deepvlarchitecture}
\end{figure}

\subsection{Network Architecture} Motivated by the light-weight architecture of \ac{gru}~\cite{cho-etal-2014-learning}, their ability to process the latest input as it arrives and propagate the temporal contexts using hidden states, we use it as the core temporal recurrent backbone. The network, outlined in Figure~\ref{fig:deepvlarchitecture},
takes the current propriocecptive vector $\mathbf{p}_{t}$ as an input of dimension $7+J$ channels to the first layer of the network, followed by 3 layers of the \acp{gru} with hidden layer dimension of $40$. The output of the last \ac{gru} layer is passed through $50\%$ dropout followed by two fully connected layers each of dimension of $40\times3$ to obtain the $3D$ outputs of the predicted robot-centric velocity $\velpred$ and corresponding uncertainty $\velunc$. Let $\theta$ be the model parameters, and $\mu_{\theta}(\cdot)$ denote the network as a function, then the model at time $t$ can be defined as:

\begin{equation}
    \velpred_{t}, \velunc_{t}, \mathbf{h}_{t}=\mu_{\theta}\left(\mathbf{p}_{t}, \mathbf{h}_{t-1}\right)
\end{equation}
where $\mathbf{h}_{t-1}$ is the \ac{gru} hidden state from the last time step $(t-1)$. Similar to TLIO \cite{liuTLIOTightLearned2020a} we use \ac{mse} to train the network in the beginning until the predictions stabilize and then use \ac{gnll} to further supervise the predictions and uncertainty. The \ac{mse} loss for a batch of size $n$ is defined as:

\begin{equation}
    \mathcal{L}_{\mathrm{\ac{mse}}}(\vel,\velpred)=\frac{1}{n}\sum_{i=1}^{n}\|\vel_{i} - \velpred_{i}\|^{2}
\end{equation}
where $\vel=\{\vel_{i}\}_{i\leq{n}}$ are the robot-centric linear velocity used as supervision. Further, the \ac{gnll} is defined as:

\begin{equation}
    \mathcal{L}_{\mathrm{\ac{gnll}}}(\vel, \sigmapred, \velpred)=\frac{1}{n}\sum_{i=1}^{n}\left(\frac{1}{2}\log{|{\sigmapred_{i}}|}+\frac{1}{2}\|\vel_{i}-\velpred_{i}\|^{2}_{\sigmapred_{i}}\right)
\end{equation}
 where $\sigmapred=\{\sigmapred_{i}\}_{i\leq{n}}$ is the covariance matrix for the term. Similar to TLIO \cite{liuTLIOTightLearned2020a} we assume a diagonal covariance matrix and define $\sigmapred\left(\velpred\right)=\textrm{diag}\left(e^{2\velunc_{x}}, e^{2\velunc_{y}}, e^{2\velunc_{z}}\right)$. Since the supervision velocity is in the frame $\bodyframe$, the principle axes of the predicted covariance is along the \ac{imu} axes.
 
\subsection{Ensemble Predictive Uncertainty} The light-weight nature of the proposed model allows to use an ensemble of models to further enhance the predictions before integrating them into state estimation. We use the predictive uncertainty as described in \cite{lakshminarayananSimpleScalablePredictive2017} based on an ensemble of the neural network models. Hence, for an ensemble of $M$ networks, with $\theta_{m}$ as the parameters, the output is a mixture of Gaussians $\mathrm{M}^{-1}\Sigma\mathcal{N}\left(\velpred_{m}, \sigmapred_{m}\right)$ with mean as:
\begin{equation}
    \velpred_{*}=\frac{1}{\mathrm{M}}\sum_{m=1}^{M}\velpred_{m}
\end{equation}
 and the covariance as:
 \begin{equation}
     \sigmapred_{*}=\frac{1}{\mathrm{M}}\sum_{m=1}^{M}\left(\sigmapred_{m}+\velpred_{m}^{2}\right)-\velpred_{*}^{2}.
 \end{equation}

To incorporate the uncertainty, we train the ensemble of $\mathrm{M}$ networks and $\velpred_{*}$ is used as the velocity update in the \ac{ekf} with $\sigmapred_{*}$ as the covariance.

\subsection{Network Implementation Details}
The network contains $28$k trainable parameters. It is trained on sequences each having a length of $15$ seconds (i.e. $300$ data points) with a batch size of $128$. A total of $\approx120$k sequences are randomly sampled from the collected data to form the training set and $\approx10$k distinct sequences (not in the training data) are used as a validation set. The Adam optimizer is used with a learning rate of $0.001$, along with a multi-step rate scheduler at $1500$, $2500$, $3500$ with gamma of $0.2$. The training is started with \ac{mse} loss, while after $3000$ iterations we switch the loss function to \ac{gnll} and the training is stopped at $4000$ iterations. Furthermore, for the ensemble predictive uncertainty, we use an ensemble of $M=8$ networks. All input and outputs other than for uncertainty are normalized to achieve 0 mean and 1.0 standard deviation. The total training takes an hour to train the network on an NVIDIA RTX $3080$ GPU. The network can run both on GPU and CPU with an inference time \SI{<5}{\milli\second} on an Orin AGX. This efficient result is attributed to the single input, single output implementation where the \ac{gru} hidden states propagate the temporal context.


\section{Integration into Inertial Odometry with optional Visual Fusion}\label{sec:vio}
We use \ac{reaqrovio} ~\cite{SinghRCMinRovio2024} (an underwater \ac{vio} method based on ROVIO~\cite{bloesch2015robust}) as the underlying \ac{ekf} based state estimation framework. We define the state as in \ac{reaqrovio}:

\small
\begin{equation}
    \mathbf{s}=\left[\mathbf{r}, \mathbf{q}, \boldsymbol{\upsilon}, \mathbf{b}_{\mathbf{a}}, \mathbf{b}_{\boldsymbol{\omega}},| \mathbf{c}, \mathbf{z},|n,| \mu_{0}, \mu_{1}, ...,\mu_{N},| \rho_{0}, \rho_{1}, ..., \rho_{N}\right]
\end{equation}
\normalsize
where $\mathbf{r}$ denotes the robot-centric position and $\boldsymbol{\upsilon}$ is the robot-centric velocity, $\mathbf{q}$ denotes the orientation of the robot as a map from the body frame $\mathcal{B}$ to the world frame $\mathcal{W}$, the $\ac{imu}$ acceleration biases are defined by $\mathbf{b}_{\mathbf{a}}$ and gyroscope biases are defined by $\mathbf{b}_{\boldsymbol{\omega}}$. The camera to \ac{imu} extrinsics are denoted by $\mathbf{c}$ for linear translation and $\mathbf{z}$ for the rotation. The refractive index of the medium is denoted by $n$. The terms $\mu_{n}$ and $\rho_{n}$ denote the bearing vector and the inverse depth of the visual features respectively. The value $N$ denotes the number of maximum features in the state of \ac{reaqrovio}.

\subsection{Velocity Update} The velocity predictions $\velpred_{*}$ and the corresponding covariance $\sigmapred_{*}$ from the proposed network are used in a newly introduced innovation term $\mathbf{y}_{\boldsymbol{\upsilon}}$ for velocity update:

\begin{equation}
    \boldsymbol{y}_{\boldsymbol{\upsilon}}= \velpred_{*} - \boldsymbol{\upsilon} + \mathbf{n}_{{\boldsymbol{\upsilon}}},\quad\mathbf{n}_{\boldsymbol{\upsilon}}\sim\mathcal{N}(\mathbf{0}, \sigmapred_*).
\end{equation}

\subsection{Relative Depth Update} Let $d_{t}$ be the depth measured by the barometric depth sensor at a given time $t$ along the negative $Z$ axis of the world frame $\mathcal{W}$. Then we use the relative depth measurement $d_{\Delta}=d_{t}-d_{0}$ (with noise variance $\sigma_{d_\Delta}^{2}$) where $d_{0}$ denotes the initial depth measured at $t=0$. The innovation term $y_{Z}$ for the $Z$ component of $\mathbf{r}$ takes the form: 

\begin{equation}
    y_{Z} = -d_{\Delta}-\mathbf{r}_{Z} + n_{d_{\Delta}},\quad n_{d_{\Delta}}\sim\mathcal{N}(0, \sigma_{d_\Delta}^2).
\end{equation}

\subsection{Contribution of Visual Features and Relative Depth}
The present work fuses the velocity from \ac{deepvl} and optionally fuses the visual patch features as described in \ac{reaqrovio}. We particularly vary the number of features $N$ in the state to emulate a presence of critically low visual features from the environment. In the remainder of the work $\ac{deepvl}_{0}$ (or $\ac{deepvl}$) refers to the fusion of velocity $\velpred_{*}$ and it covariance $\sigmapred_{*}$ from the presented network in \ac{reaqrovio} with relative depth update and with no visual features in the state $\mathbf{s}$. Whereas, $\ac{deepvl}_{N}$ denotes a framework with additional inclusion of $N$ visual features in the state. Similarly, for concise notation of the various configurations of \ac{reaqrovio}, we denote $\ac{vio}_{N}$ as the a visual-inertial odometry framework from \ac{reaqrovio} with $N$ features in the state and with relative depth update.

\begin{figure*}[ht!]
\centering
    \includegraphics[width=0.98\textwidth]{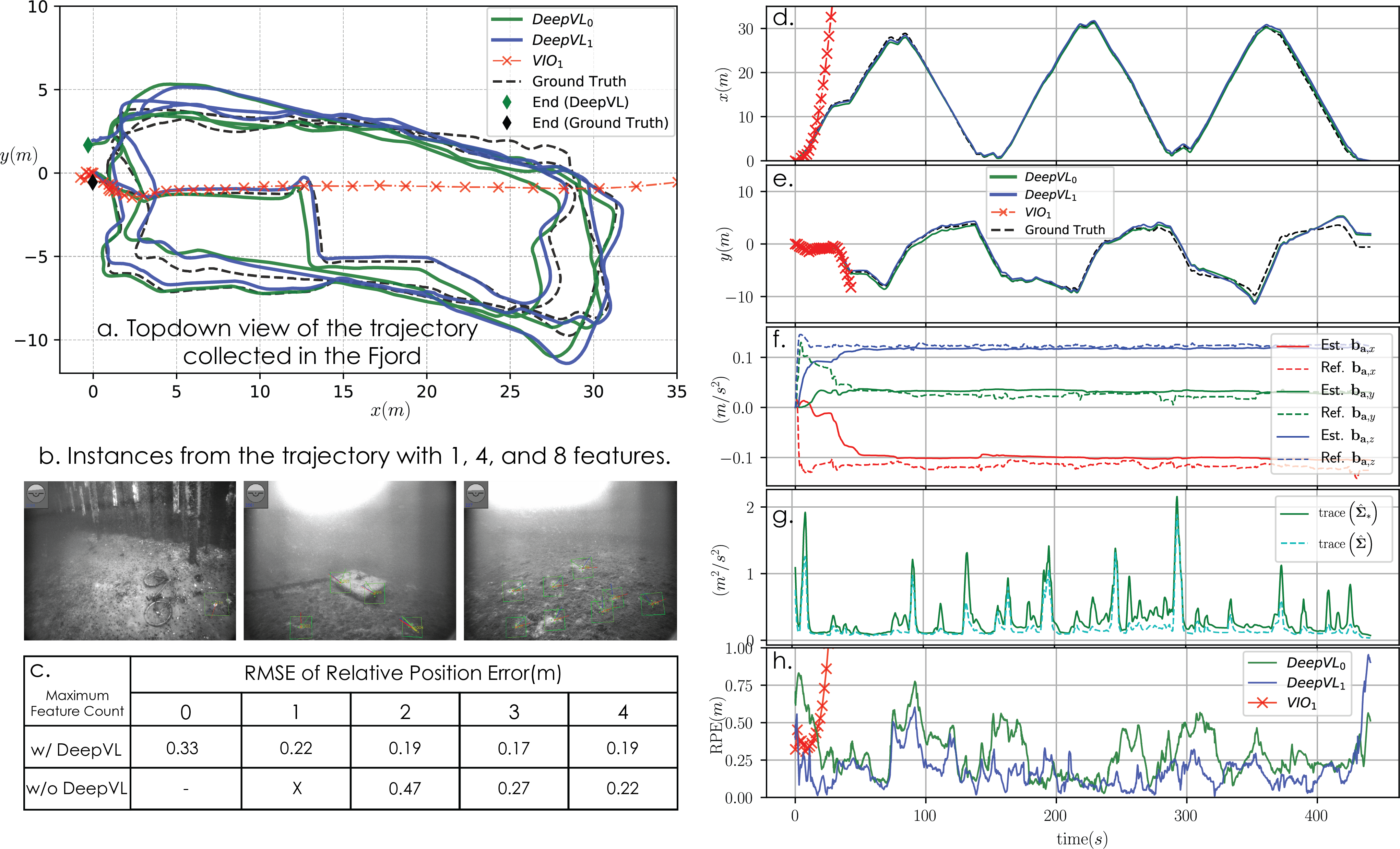}
\vspace{-2ex} 
\caption{Detailed analysis of trajectory $5$ collected in the Trondheim Fjord. a) The odometry estimate with \ac{deepvl}, \ac{vio} with $1$ feature, and fusion of \ac{deepvl} with \ac{vio} with $1$ feature. b) Images from the Alphasense camera stream from multiple locations in the trajectory. c) Tabular comparison of \ac{rpe} with maximum features ranging from $0$ to $8$ (`X' indicating divergence, while `-' indicates that a test is not ran if not meaningful). On the right, the evolution of position, accelerometer biases, the uncertainty estimates and \ac{rpe} are shown.}
\label{fig:detailed_results}
\vspace{-2ex} 
\end{figure*}

\begin{figure}
    \centering
    \includegraphics[width=1\linewidth]{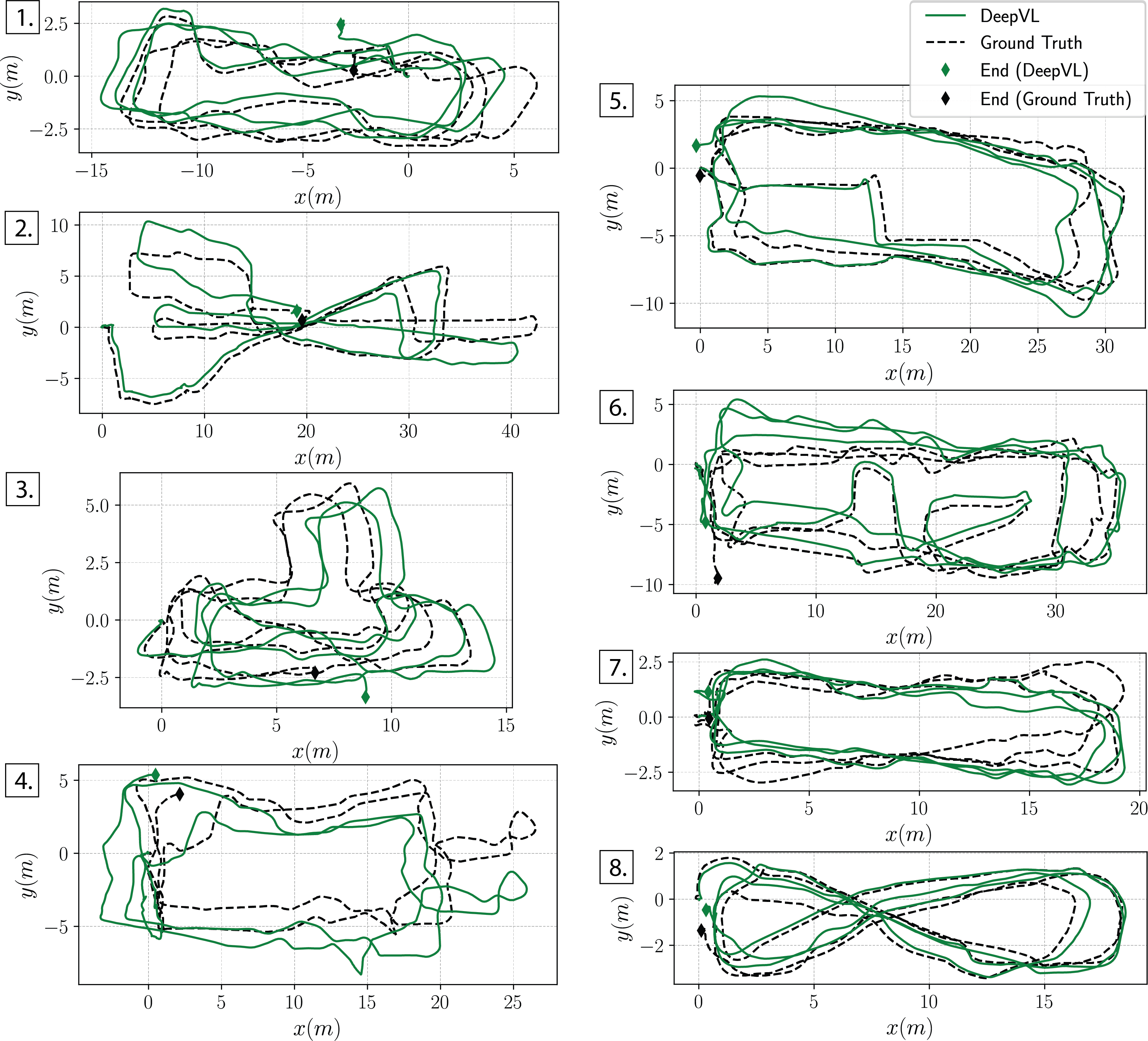}
    \caption{A collective plot showcasing all the $8$ evaluation trajectories along with odometry estimates based on \ac{deepvl}.}
    \label{fig:collective_plot}
\end{figure}

\section{Evaluation Studies}\label{sec:evaluation}
\subsection{Training Dataset Collection}
We used the BlueROV \ac{rov} further integrating an AlphaSense Core Research 5 Camera-\ac{imu} module to collect the data for training. The Alphasense module integrates the BMI085 \ac{imu} running at \SI{200}{\hertz}, and $5$ Sony IMX-287 cameras running at \SI{20}{\hertz}. We use a Pixhawk6X as the autopilot and Orin AGX \SI{32}{\giga\byte} as the onboard computer. The motor commands and the battery percentage are measured at \SI{20}{\hertz} from the autopilot via the MAVLink protocol. The Alphasense module is synchronized with the Orin AGX computing system via PTP, while the autopilot is connected via Ethernet and runs time synchronisation via MAVROS. To maximize the performance of the reference odometry, we use 4 of the cameras (the fifth is looking upwards and thus not used), namely the front-facing stereo pair (inclined by \SI{16}{\degree}) and the left and right side facing monocular cameras. \ac{reaqrovio}~\cite{SinghRCMinRovio2024} with $N=30$ features ($\ac{vio}_{30}$) is used to estimate the odometry and the robot-centric velocity of the vehicle, considering the data from all four cameras and the \ac{imu}, and its output is treated as ground-truth to supervise the network's training and assess the performance of the method in the evaluations presented further in this section. The intrinsics and extrinsics calibration for all cameras is performed in air and the refractive camera model in \cite{SinghRCMinRovio2024} adapts the cameras to the water by using a fixed value for the refractive index i.e. $1.33$ for the freshwater environments and $1.34$ for seawater environments respectively. 

We collected data in the laboratory pool of NTNU's Marine Cybernetics lab (MC-lab) and in the Trondheim Fjord. The dimensions of the MC-lab pool are $\SI{40}{\meter}\times\SI{6.45}{\meter}\times\SI{1.5}{\meter}$, while the average depth of the experiment site in the Fjord is \SI{6}{\meter}. The robot is piloted manually to perform a diverse set of motions in order to cover most of the state space, and such that both low- and high-frequency motions are experienced. Particularly, the trajectories involve velocities uniformly ranging from $0$ to \SI{\approx 0.8}{\meter/\second}, in all linear directions. The majority of the data was collected in the pool of MC-lab, which is about 1.5 meters deep thus extended vertical motions could not be captured. The dataset includes 3 hours and 40 minutes from the MC-lab and 20 minutes in the Trondheim Fjord, with the latter allowing for extended variations in $Z$ and also including long-duration static motions. In the MC-lab, the robot was piloted to move in all lateral directions for varying amounts, and the motion patterns were varied at random instances. Overall the key insight in data collection is to acquire a representative sample set of the motions a robot can undergo with the feasible limits. To further extend the envelope, we also change the attitude of the robot to random orientations using the ``Attitude Hold'' mode of the onboard autopilot. The robot was connected to the ground-station computer via a tether cable which is used for high-level communication and telemetry. It was ensured that the tether was tension-free to avoid any unaccounted force on the robot. The collected data shall be releasd in \url{https://github.com/ntnu-arl/underwater-datasets}. 

\subsection{Evaluation Dataset} For the evaluation of the proposed method we collected $6$ trajectories of varying lengths and coverage in the Fjord as showcased in Figure \ref{fig:collective_plot} (trajectory $1$ to $6$). Additionally, we collect $2$ trajectories ($7$ and $8$) in the indoor pool (MC-lab).

\subsection{Detailed Evaluation on a representative trajectory}
We present a detailed result on the trajectory $5$ collected in the Fjord. The length of the trajectory is \SI{235}{\meter} and the maximum linear velocity of the robot is \SI{0.72}{\meter/\second} while the average velocity is \SI{0.52}{\meter/\second}. Its total duration is \SI{440}{\second}, and it consists of three laps starting from the origin and reaching the endpoint as shown in Figure \ref{fig:detailed_results}(a). First, the robot descends from the surface, and moves along the wall, while it is then piloted to move along the seabed overall ensuring that in general there are visual features such that the result of \ac{reaqrovio} with all $4$ cameras ($\ac{vio}_{30}$) can be considered a reasonable ground-truth. The executed trajectory is shown in Figure \ref{fig:detailed_results}. We compare the odometry estimates from $\ac{deepvl}_{0}$ (i.e. with no camera feed) and show that it achieves a Relative Position Error (\ac{rpe}) \ac{rmse} of \SI{0.33}{\meter} over deltas of \SI{10}{\meter} (\SI{\approx3}{\percent}). We also present the result for $\ac{deepvl}_{1}$  from a monocular camera against $\ac{vio}_{1}$. In the present trajectory, the estimates of $\ac{vio}_{1}$ diverges. We also compare the \ac{imu} biases estimated by $\ac{deepvl}_{0}$ over time against the reference ground-truth odometry estimation and the bias convergence is verified. Furthermore, we compare the trace of uncertainty predicted by the network $\sigmapred$ and predicted from the ensemble of the networks $\sigmapred_{*}$. The former is consistently overconfident about the velocity predictions. The effect of network uncertainty is further discussed in Section~\ref{subsection:ablation}. Last, we present the \ac{rpe} over time which also validates the enhancement in the odometry performance from $\ac{deepvl}_{0}$ to $\ac{deepvl}_{1}$.
\begin{figure}
    \centering
    \includegraphics[width=1\linewidth]{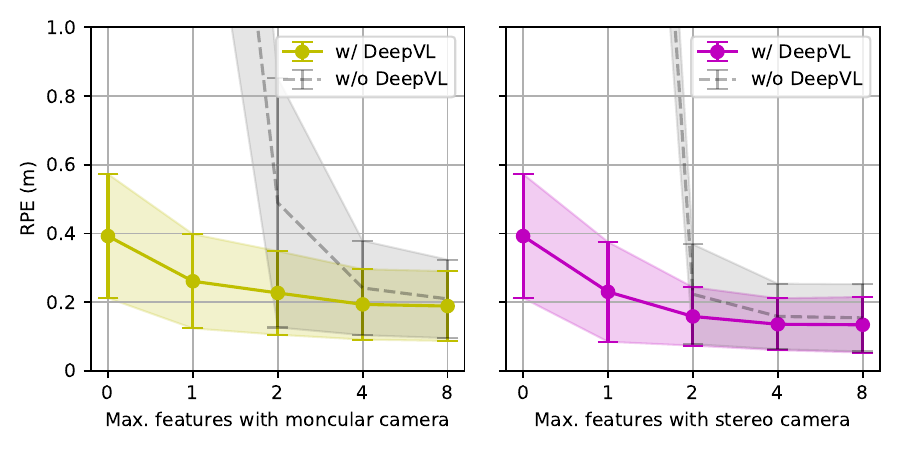}
    \caption{Aggregate \ac{rpe} over all trajectories with monocular (left) and stereo (right) camera to analyze the effect of incrementally increasing the number of maximum features used in \ac{vio} with and without the integration of \ac{deepvl}.}
    \label{fig:aggregate_rpe}
\end{figure}
\subsection{Collective evaluation on all trajectories}
We further evaluate the proposed method on the full dataset of $8$ trajectories (Figure~\ref{fig:collective_plot}). Firstly, we evaluate $\ac{deepvl}_{0}$ and the average $\ac{rpe}$ over all trajectories over deltas of \SI{10}{\meter} is \SI{0.39}{\meter}. We further vary the number of features $N$ ranging from $(1, 2, 4, 8)$ in both $\ac{vio}_{N}$ and $\ac{deepvl}_{N}$. Figure \ref{fig:aggregate_rpe} shows the plot of \ac{rpe} values and corresponding standard deviation for $\ac{deepvl}_{N}$ and $\ac{vio}_{N}$ for both the front-left monocular and the stereo pair of the Alphasense camera. Likewise, Table~\ref{table:average_rpe_mono_stereo} summarizes \ac{rpe} values across the dataset. As shown, the \ac{rpe} for $\ac{deepvl}_{N}$ is consistently lower than $\ac{vio}_{N}$ for the same number of features.

\begin{table}[]
\caption{Table presenting the average of the \ac{rpe} for varying visual features for monocular and stereo camera configuration. The symbol 'X' indicates that a method diverges. The \ac{rpe} was calculated over delta of \SI{10}{\meter}.}
\resizebox{\columnwidth}{!}{%
\begin{tabular}{|l|lllll|}
\hline
             & \multicolumn{5}{l|}{Maximum Visual Features (N)}                                                     \\ \hline
\textbf{}    & \multicolumn{1}{l|}{0} & \multicolumn{1}{l|}{1}           & \multicolumn{1}{l|}{2}           & \multicolumn{1}{l|}{3}           & 4           \\ \hline
$\ac{deepvl}_{N}$(Mono) & \multicolumn{1}{l|}{0.393} & \multicolumn{1}{l|}{0.260} & \multicolumn{1}{l|}{0.226} & \multicolumn{1}{l|}{0.193} & 0.188 \\ \hline
$\ac{vio}_{N}$ (Mono)& \multicolumn{1}{l|}{-} & \multicolumn{1}{l|}{X} & \multicolumn{1}{l|}{0.489} & \multicolumn{1}{l|}{0.241} & 0.209  \\ \hline
$\ac{deepvl}_{N}$(Stereo)  & \multicolumn{1}{l|}{0.393} & \multicolumn{1}{l|}{0.229} & \multicolumn{1}{l|}{0.158} & \multicolumn{1}{l|}{0.135} & 0.134 \\ \hline
$\ac{vio}_{N}$  (Stereo)& \multicolumn{1}{l|}{-} & \multicolumn{1}{l|}{X} & \multicolumn{1}{l|}{0.222} & \multicolumn{1}{l|}{0.158} & 0.154 \\ \hline
\end{tabular}%
}\label{table:average_rpe_mono_stereo}
\end{table}

\begin{figure}
    \centering
    \includegraphics[width=1\linewidth]{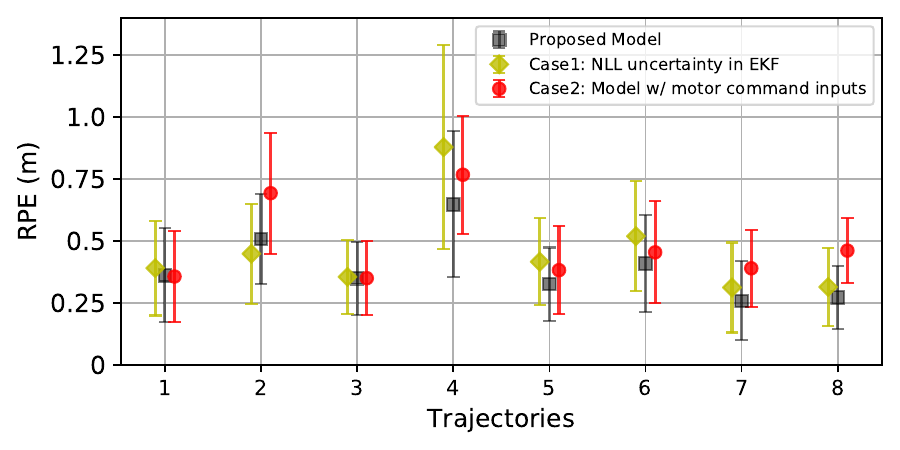}
    \caption{Ablation results for comparison between the odometry results from the proposed network, the proposed network with \ac{gnll} uncertainty and a network that only uses motor commands as inputs.} 
    \label{fig:ablation}
\end{figure}

\subsection{Ablation Study}\label{subsection:ablation}
To evaluate the proposed model, we conduct the following ablations: `Case$1$': We first replace the proposed ensemble-based predictive uncertainty by a single model \ac{gnll}-based uncertainty. It is observed that the former achieves superior \ac{rpe} scores for odometry compared to the latter. `Case$2$': We evaluate the network that is only trained on the motor command data as input to analyse the contribution of the motor commands in the proposed network. Figure \ref{fig:ablation} shows the \ac{rpe} for both the above cases and the proposed model for every trajectory, while Table \ref{table:ablation} shows the average \ac{rpe} for all trajectories. The proposed  model with the \ac{imu}, motor commands and battery inputs performs superior compared to the model trained only on motor inputs. Similarly, we train a model only with \ac{imu} measurements as the input and evaluate the resulting odometry performance where the average (over all $8$ trajectories) \ac{rpe} is \SI{3.90}{\meter} compared to \SI{0.39}{\meter} for the proposed network, and \SI{0.48}{\meter} for the network trained only with motor commands. These ablations verify the use of the ensemble-based approach, and provide insight into the complementary nature of \ac{imu} measurements and motor command inputs to the model.

\subsection{Closed loop position control verification}
\begin{figure}
    \centering
    \includegraphics[width=1\linewidth]{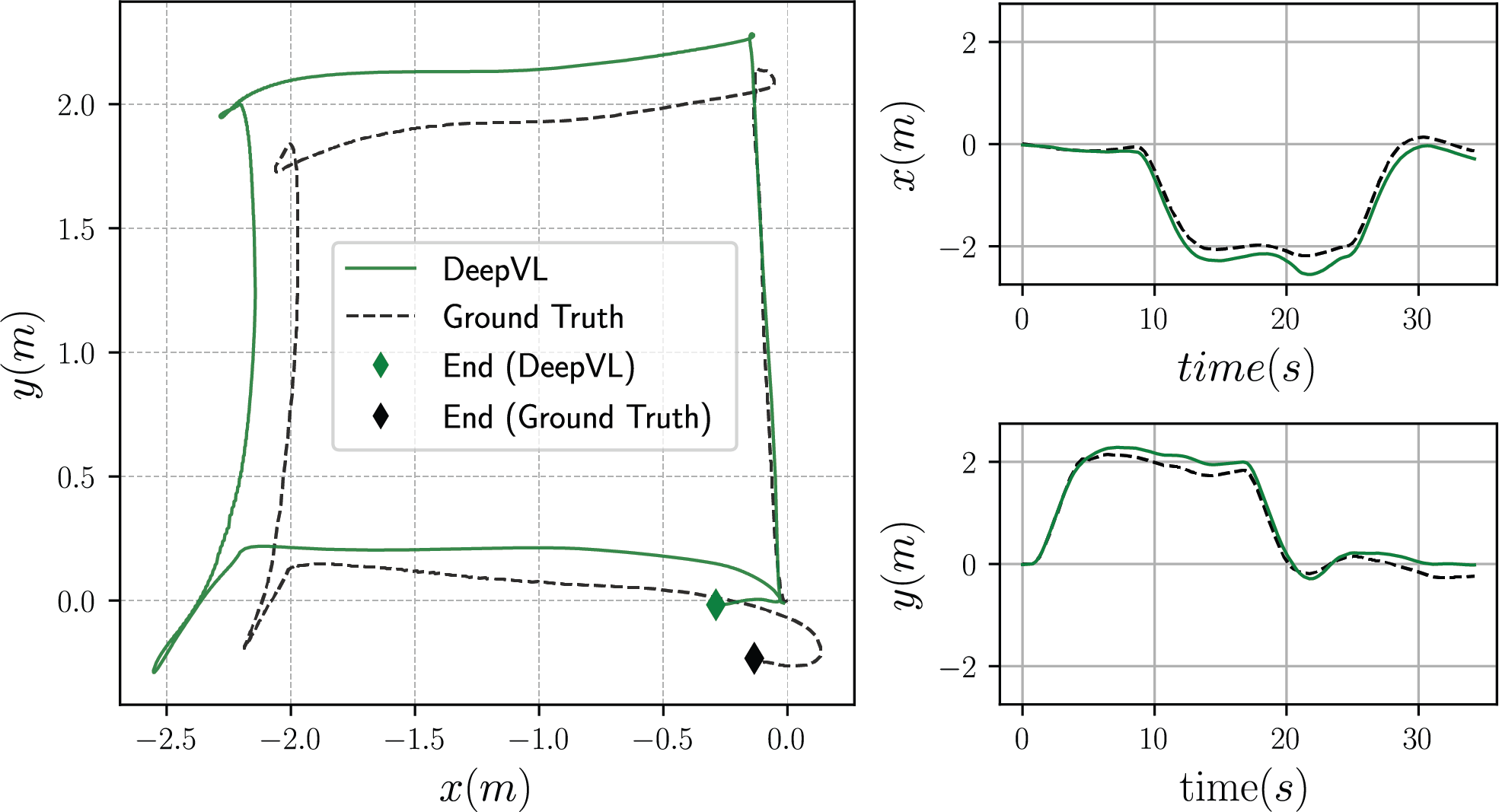}
    \caption{The plot on the left showing topdown view of the online estimated position with \ac{deepvl} running onboard the robot which is used for following position set-points along the vertices of a square of edge length \SI{2}{\meter}. The two plots on right show the estimated position vs. time. The ``Ground-Truth'' is estimated offline using \ac{vio}.}
    \label{fig:position_control}
\end{figure}

We use the presented method to perform state estimation onboard the robot in the MC-lab and command it to follow a \SI{2}{\meter}-wide square path through closed-loop control (Figure~\ref{fig:position_control}). The state estimation does not include any visual features and instead relies exclusively on \ac{imu}, motor data, battery readings, and barometer measurements through \ac{deepvl} as in Sections~\ref{sec:deepvl},~\ref{sec:vio}. This experiment demonstrates that the proposed method a) allows robot autonomy to perform at a functional level in the case of complete loss of vision and b) generalizes on motions that have not been in the training data. Specifically, here the network experiences closed-loop position control data which typically contain different frequency content compared to human steering. 

\begin{table}[]
\caption{Table presenting the average of the \ac{rpe} for the ablation of the network.}
\begin{tabular}{|l|l|l|l|l|}
\hline
\textbf{RPE}    & Proposed & Case1  & Case2 & IMU Only \\ \hline
\textbf{Avg RMSE (m)} & 0.393        & 0.454       & 0.482      & 3.907    \\ \hline
\textbf{Avg STD (m)}  & 0.181        & 0.211       & 0.185      & 2.031    \\ \hline
\end{tabular}%
\label{table:ablation}
\end{table}

\section{CONCLUSIONS}\label{sec:concl}
This work presented \ac{deepvl}, a Dynamics and Inertial-based method to predict velocity and uncertainty which is fused into an EKF along with a barometer to perform long-term underwater robot odometry in lack of extroceptive constraints. Evaluated on data from the Trondheim Fjord and a laboratory pool, the method achieves an average of \SI{4}{\percent} RMSE RPE compared to a reference trajectory from \ac{reaqrovio} with $30$ features and $4$ Cameras. The network contains only $28$K parameters and runs on both GPU and CPU in \SI{<5}{\milli\second}. While its fusion into state estimation can benefit all sensor modalities, we specifically evaluate it for the task of fusion with vision subject to critically low numbers of features. Lastly, we also demonstrated position control based on odometry from \ac{deepvl}.

\addtolength{\textheight}{-8cm}   

\bibliographystyle{IEEEtran}
\bibliography{BIB/main, BIB/learnedInertial}

\end{document}